%% file: main.tex
\documentclass[10pt,twocolumn,letterpaper]{article}
\pdfoutput=1
\usepackage{cvpr}
\usepackage{times}
\usepackage{epsfig}
\usepackage{graphicx}
\usepackage{amsmath}
\usepackage{amssymb}
\usepackage{booktabs}
\usepackage{authblk}
\usepackage{pifont}
\newcommand{\cmark}{\ding{51}}%
\newcommand{\xmark}{\ding{55}}%
\usepackage{paralist, tabularx}
\usepackage{authblk}

\makeatletter
\renewcommand\AB@affilsepx{, \protect\Affilfont}
\makeatother

\usepackage{cite}

\cvprfinalcopy 

\def\cvprPaperID{2722} 

\ifcvprfinal\fi
\begin{document}

\title{RVOS: End-to-End Recurrent Network for Video Object Segmentation}

\author{Carles Ventura$^1$, Miriam Bellver$^2$, Andreu Girbau$^3$, Amaia Salvador$^3$,
\\ \vspace{-3mm}
Ferran Marques$^3$ and Xavier Giro-i-Nieto$^3$
\\ \vspace{3mm}
$^1$Universitat Oberta de Catalunya \:\:\:
$^2$Barcelona Supercomputing Center \\
$^3$Universitat Polit\`ecnica de Catalunya \\ \vspace{2mm}
\small{cventuraroy@uoc.edu, miriam.bellver@bsc.es, \{andreu.girbau, amaia.salvador, ferran.marques, xavier.giro\}@upc.edu} }



\maketitle

\input{sections/0_abstract.tex}
\input{sections/1_introduction.tex}

\input{sections/2_related.tex}
\input{sections/3_model.tex}
\input{sections/4_experiments.tex}
\input{sections/5_conclusion.tex}
\input{sections/6_acks.tex}

\input{main.bbl}
\end{document}

%% file: sections/0_abstract.tex
\begin{abstract}

Multiple object video object segmentation is a challenging task, specially for the zero-shot case, when no object mask is given at the initial frame and the model has to find the objects to be segmented along the sequence.
In our work, we propose a Recurrent network for multiple object Video Object Segmentation (RVOS) that is fully end-to-end trainable. Our model incorporates recurrence on two different domains: $(i)$ the spatial, which allows to discover the different object instances within a frame, and $(ii)$ the temporal, which allows to keep the coherence of the segmented objects along time. We train RVOS for zero-shot video object segmentation and are the first ones to report quantitative results for DAVIS-2017 and YouTube-VOS benchmarks. Further, we adapt RVOS for one-shot video object segmentation by using the masks obtained in previous time steps as inputs to be processed by the recurrent module. Our model reaches comparable results to state-of-the-art techniques in YouTube-VOS benchmark and outperforms all previous video object segmentation methods not using online learning in the DAVIS-2017 benchmark. Moreover, our model achieves faster inference runtimes than previous methods, reaching 44ms/frame on a P100 GPU. 
\end{abstract}

%% file: sections/1_introduction.tex
\section{Introduction}
\label{sec:intro}

Video object segmentation (VOS) aims at separating the foreground from the background given a video sequence. This task has raised a lot of interest in the computer vision community since the appearance of benchmarks~\cite{Perazzi2016} that have given access to annotated datasets and standardized metrics. Recently, new benchmarks~\cite{Pont-Tuset_arXiv_2017, xu2018youtube-benchmark} that address multi-object segmentation and provide larger datasets have become available, leading to more challenging tasks.

\begin{figure}
  \centering
  \includegraphics[width=\columnwidth]{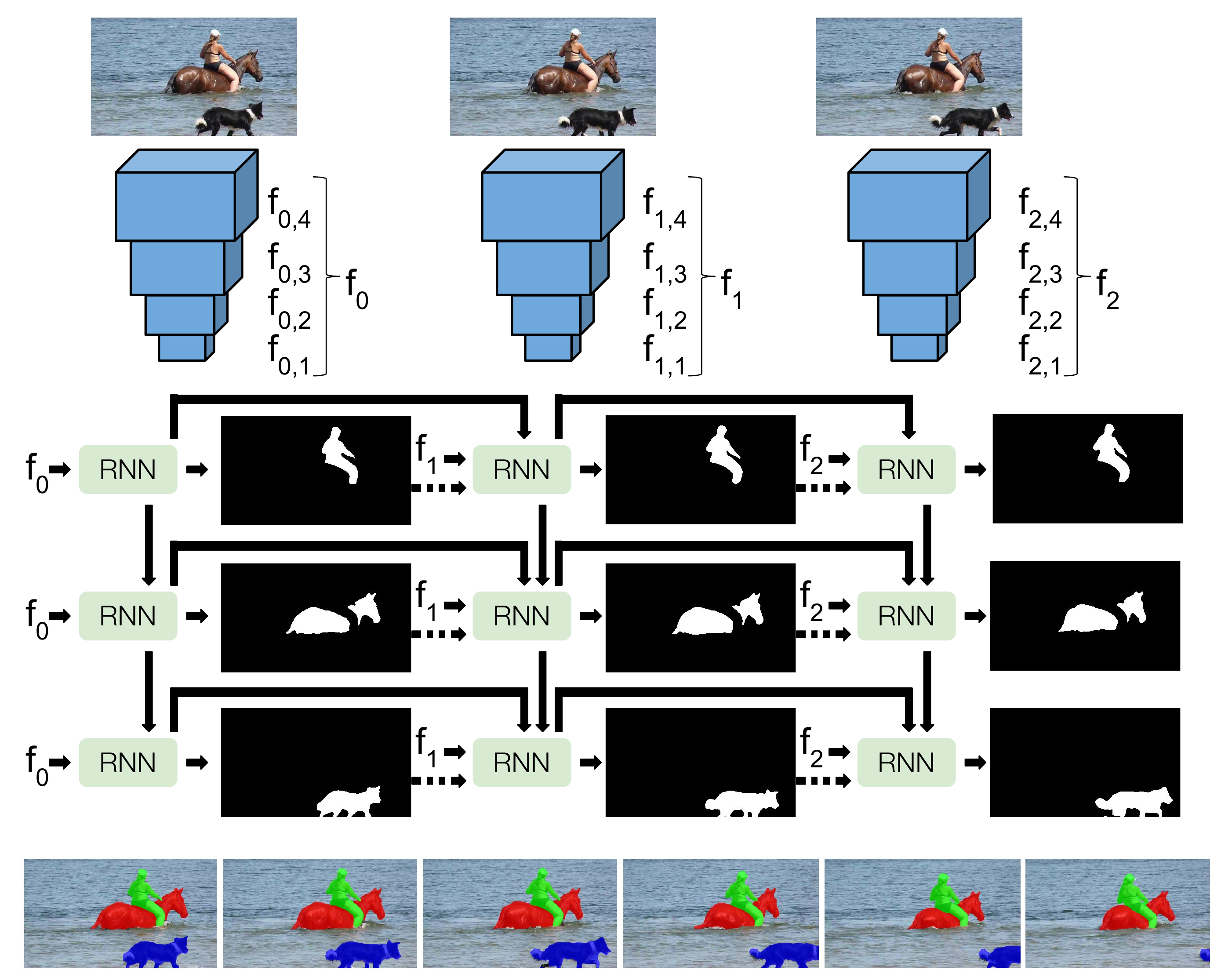}
  \caption{Our proposed architecture where RNN is considered in both spatial and temporal domains. We also show some qualitative results where each predicted instance mask is displayed with a different color.}
  \label{fig:spatiotemporal-RSIS}
\end{figure}

Most works addressing VOS treat frames independently~\cite{caelles2017one, voigtlaender2017online, maninis2018video, Cheng_favos_2018}, and do not consider the temporal dimension to gain coherence between consecutive frames. Some works have leveraged the temporal information using optical flow estimations~\cite{cheng2017segflow, jain2017fusionseg, tokmakov2017learning, bao2018cnn} or propagating the predicted masks through the video sequence~\cite{perazzi2017learning, yang2018efficient}.



In contrast to these works, some methods propose to train models on 
spatio-temporal features, e.g., \cite{tokmakov2017learning} used RNNs to encode the spatio-temporal evolution of objects in the video sequence. However, their pipeline relies on an optical flow stream that prevents a fully end-to-end trainable model. 
Recently, \cite{xu2018youtube} proposed an encoder-decoder architecture based on RNNs that is similar to our proposed pipeline. The main difference is that they process only a single object in an end-to-end manner. Thus, a separate forward pass of the model is required for each object that is present in the video. None of these models consider multi-object segmentation in a unified manner.  

We present an architecture (see Figure~\ref{fig:spatiotemporal-RSIS}) that serves for several video object segmentation scenarios (single-object vs. multi-object, and one-shot vs. zero-shot). Our model is based on RSIS \cite{Salvador17}, a recurrent model for instance segmentation that predicts a mask for each object instance of the image at each step of the recurrence. Thanks to the RNN's memory capabilities, the output of the network does not need any post-processing step since the network learns to predict a mask for each object. In our model for video object segmentation, we add recurrence in the temporal domain to predict instances for each frame of the sequence.

The fact that our proposed method is recurrent in the spatial~(the different instances of a single frame) and the temporal~(different frames) domains allows that the matching between instances at different frames can be handled naturally by the network. For the spatial recurrence, we force that the ordering in which multiple instances are predicted is the same across temporal time steps. Thus, our model is a fully end-to-end solution, as we obtain multi-object segmentation for video sequences without any post-processing. 


Our architecture addresses the challenging task of zero-shot learning for VOS (also known as unsupervised VOS in a new challenge from DAVIS-2019\footnote{{\href{https://davischallenge.org/challenge2019/unsupervised.html}{https://davischallenge.org/challenge2019/unsupervised.html}}}). In this case, no initial masks are given, and the model should discover segments along the sequences. We present quantitative results for zero-shot learning for two  benchmarks: DAVIS-2017~\cite{Pont-Tuset_arXiv_2017} and YouTube-VOS~\cite{xu2018youtube-benchmark}. Furthermore, we can easily adapt our architecture for one-shot VOS (also known as semi-supervised), by feeding the objects masks from previous time steps to the input of the recurrent network. 
Our contributions can be summarized as follows: 
\begin{itemize}
    \item We present the first end-to-end architecture for video object segmentation that tackles multi-object segmentation and does not need any post-processing.
    \item Our model can easily be adapted to one-shot and zero-shot scenarios, and we present the first quantitative results for zero-shot video object segmentation for the DAVIS-2017 and Youtube-VOS benchmarks~\cite{Pont-Tuset_arXiv_2017, xu2018youtube-benchmark}.
    \item We outperform previous VOS methods which do not use online learning. Our model achieves a remarkable performance without needing finetuning for each test sequence,  becoming the fastest method.
\end{itemize}


%% file: sections/2_related.tex
\section{Related Work}
Deep learning techniques for the object segmentation task have gained attention in the research community during the recent years \cite{caelles2017one, voigtlaender2017online, perazzi2017learning, yang2018efficient, cheng2017segflow, jampani2017video, tokmakov2017learning, hu2017maskrnn, song2018pyramid, koh2017primary, tokmakov2017learning_motion, jain2017fusionseg, lao2018extending, hu2018unsupervised, xingjian2015convolutional, Salvador17}. In great measure, this is due to the emergence of new challenges and segmentation datasets, from Berkeley Video Segmentation Dataset (2011) \cite{amfm_pami2011}, SegTrack (2013) \cite{li2013video}, Freiburg-Berkeley Motion Segmentation Dataset (2014) \cite{ochs2014segmentation}, to more accurate and dense labeled ones as DAVIS (2016-2017) \cite{Perazzi2016, Pont-Tuset_arXiv_2017}, to the latest segmentation dataset YouTube-VOS (2018) \cite{xu2018youtube}, which provides the largest amount of annotated videos up to date.

\textbf{Video object segmentation}
Considering the temporal dimension of video sequences, we differentiate between algorithms that aim to model the temporal dimension of an object segmentation through a video sequence, and those without temporal modeling that predict object segmentations at each frame independently. 

For segmentation without temporal modeling, one-shot VOS has been handled with online learning, where the first annotated frame of the video sequence is used to fine-tune a pretrained network and segment the objects in other frames~\cite{caelles2017one}. 
Some approaches have worked on top of this idea, by either updating the network online with additional high confident predictions~\cite{voigtlaender2017online}, or by using the instance segments of the different objects in the scene as prior knowledge and blend them with the segmentation output~\cite{maninis2018video}.
Others have explored data augmentation strategies for video by applying transformations to images and object segments~\cite{khoreva2017lucid}, tracking of object parts to obtain region-of-interest segmentation masks \cite{Cheng_favos_2018}, or meta-learning approaches to quickly adapt the network to the object mask given in the first frame~\cite{yang2018efficient}. 

To leverage the temporal information, some works~\cite{cheng2017segflow, jain2017fusionseg, tokmakov2017learning, nilsson2018} depend on pretrained models on other tasks (e.g. optical flow or motion segmentation). Subsequent works \cite{bao2018cnn} use optical flow for temporal consistency after using Markov random fields based on features taken from a Convolutional Neural Network.  
An alternative to gain temporal coherence is to use the predicted masks in the previous frames as guidance for next frames~\cite{perazzi2017learning, yang2018efficient, hu2017maskrnn, jang2017}. In the same direction, \cite{jampani2017video} propagate information forward by using spatio-temporal features.
Whereas these works cannot be trained end-to-end, we propose a model that relies on the temporal information and can be fully trained end-to-end for VOS. Finally, \cite{xu2018youtube} makes use of an encoder-decoder recurrent neural network structure, that uses Convolutional LSTMs for sequence learning. One difference between our work and \cite{xu2018youtube} is that our model is able to handle multiple objects in a single forward pass by including spatial recurrence, which allows the object being segmented to consider previously segmented objects in the same frame.

\textbf{One and zero-shot video object segmentation}
In video object segmentation, one-shot learning is understood as making use of a single annotated frame (often the first frame of the sequence) to estimate the remaining frames segmentation in the sequence. On the other hand, zero-shot or unsupervised learning is understood as building models that do not need an initialization to generate segmentation masks of objects in the video sequence. 

In the literature there are several works that rely on the first mask as input to propagate it through the sequence \cite{caelles2017one, voigtlaender2017online, perazzi2017learning, yang2018efficient, jampani2017video, tokmakov2017learning, hu2017maskrnn}. In general, one-shot methods reach better performance than zero-shot ones, as the initial segmentation is already given, thus not having to estimate the initial segmentation mask from scratch.
Most of these models rely on online learning, i.e. adapting their weights given an initial frame and its corresponding masks. Typically online learning methods reach better results, although they require more computational resources. In our case, we do not rely on any form of online learning or post-processing to generate the prediction masks.

In zero-shot learning, in order to estimate the segmentation of the objects in an image, several works have exploited object saliency \cite{song2018pyramid, jain2017fusionseg, hu2018unsupervised}, leveraged the outputs of object proposal techniques \cite{koh2017primary} or used a two-stream network to jointly train with optical flow \cite{cheng2017segflow}. Exploiting motion patterns in videos has been studied in \cite{tokmakov2017learning_motion}, while \cite{lao2018extending} formulates the inference of a 3D flattened object representation and its motion segmentation. Finally, a foreground-background segmentation based on instance embeddings has been proposed in \cite{li2018}.

Our model is able to handle both zero and one-shot cases. In Section \ref{sec:experiments} we show results for both configurations, tested on the  Youtube-VOS~\cite{xu2018youtube-benchmark} and DAVIS-2017~\cite{Pont-Tuset_arXiv_2017} datasets. For one-shot VOS our model has not been finetuned with the mask given at the first frame. Furthermore, on the zero-shot case, we do not use any pretraining on detection tasks or rely on object proposals. This way, our model can be fully trained end-to-end for VOS, without depending on models that have been trained for other tasks.

\textbf{End-to-end training} Regarding video object segmentation we distinguish between two types of end-to-end training. A first type of approach is frame-based and allows end-to-end training for multiple-objects~\cite{voigtlaender2017online, maninis2018video}. A second group of models allow training in the temporal dimension in an end-to-end manner, but deal with a single object at a time~\cite{xu2018youtube}, requiring a forward pass for each object and a post-processing step to merge the predicted instances. 

To the best of our knowledge, our model is the first that allows a full end-to-end training given a video sequence and its masks, without requiring any kind of post-processing. 

\label{sec:relatedwork}

%% file: sections/3_model.tex
\section{Model}
\label{sec:model}

We propose a model based on an encoder-decoder architecture to solve two different tasks for the video object segmentation problem: one-shot and zero-shot VOS. On the one hand, for the one-shot VOS, the input consists of the set of RGB image frames of the video sequence, as well as the masks of the objects at the frame where each object appears for first time. On the other hand, for the zero-shot VOS, the input only consists of the set of RGB image frames. In both cases, the output consists of a sequence of masks for each object in the video, with the difference that the objects to segment are unknown in the zero-shot VOS task.

\subsection{Encoder}
\begin{figure*}
  \centering
  \includegraphics[width=\textwidth]{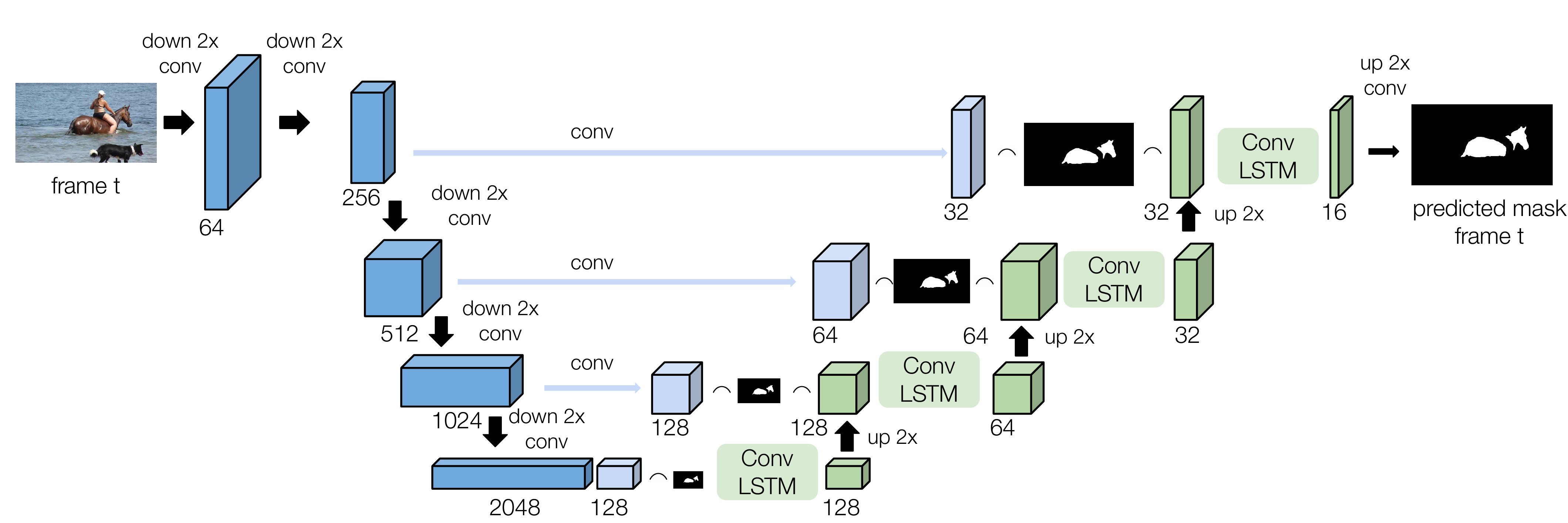}
  \caption{Our proposed recurrent architecture for video object segmentation for a a single frame at time step \textit{t}. The figure illustrates a single forward of the decoder, predicting only the first mask of the image.}
  \label{fig:encoder-decoder-mask}
\end{figure*}

We use the architecture proposed by \cite{Salvador17}, which consists of a ResNet-101~\cite{resnet} model pre-trained on ImageNet~\cite{imagenet}. This architecture does instance segmentation by predicting a sequence of masks, similarly to \cite{romera2016,ren2017}. The input $x_t$ of the encoder is an RGB image, which corresponds to frame $t$ in the video sequence, and the output $f_t = \{f_{t,1}, f_{t,2},..., f_{t,k}\}$ is a set of features at different resolutions. The architecture of the encoder is illustrated as the blue part (on the left) in Figure~\ref{fig:encoder-decoder-mask}. We propose two different configurations: $(i)$ an architecture that includes the mask of the instances from the previous frame as one additional channel of the output features (as showed in the figure), and $(ii)$ the original architecture from \cite{Salvador17}, i.e. without the additional channel. The inclusion of the mask from the previous frame is especially designed for the one-shot VOS task, where the first frame masks are given.

\subsection{Decoder}

Figure~\ref{fig:encoder-decoder-mask} depicts the decoder architecture for a single frame and a single step of the spatial recurrence. The decoder is designed as a hierarchical recurrent architecture of ConvLSTMs~\cite{xingjian2015convolutional} which can leverage the different resolutions of the input features $f_t = \{f_{t,1}, f_{t,2},..., f_{t,k}\}$, where ${f_{t,k}}$ are the features extracted at the level $k$ of the encoder for the frame $t$ of the video sequence. The output of the decoder is a set of object segmentation predictions $\{S_{t,1},,...,S_{t,i},...,S_{t,N}\}$, where $S_{t,i}$ is the segmentation of object $i$ at frame $t$. The recurrence in the temporal domain has been designed so that the mask predicted for the same object at different frames has the same index in the spatial recurrence. For this reason, the number of object segmentation predictions given by the decoder is constant ($N$) along the sequence. This way, if an object $i$ disappears in a sequence at frame $t$, the expected segmentation mask for object $i$, i.e. $S_{t,i}$, will be empty at frame $t$ and the following frames. We do not force any specific order in the spatial recurrence for the first frame. Instead, we find the optimal assignment between predicted and ground truth masks with the Hungarian algorithm using the soft Intersection over Union score as cost function.

\begin{figure}

  \centering
  \includegraphics[width=\columnwidth]{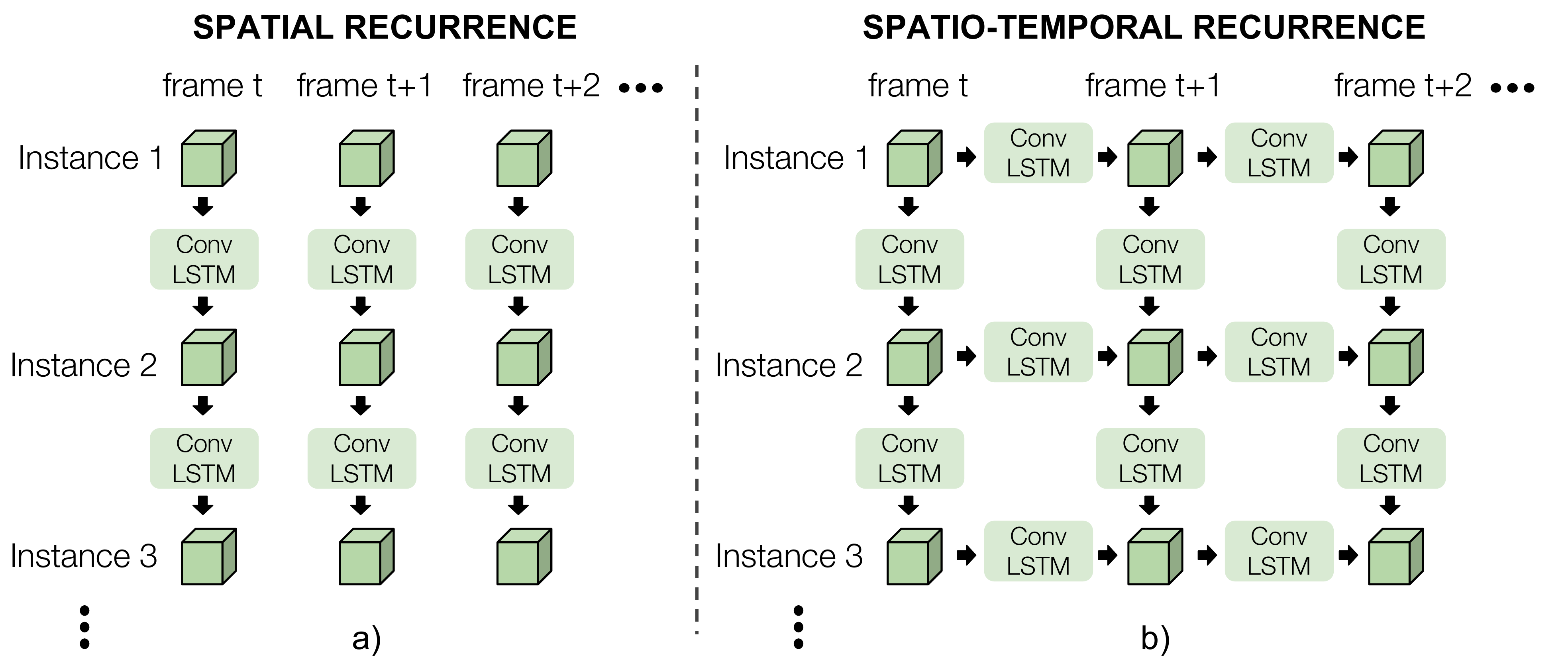}
  \caption{Comparison between original spatial \cite{Salvador17} (left) and proposed spatio-temporal recurrent networks (right).}
  \label{fig:spatiotemporal-decoder}
\end{figure}

In Figure~\ref{fig:spatiotemporal-decoder} the difference between having only spatial recurrence, over having spatial and temporal recurrence is depicted. 
The output $h_{t,i,k}$ of the $k$-th ConvLSTM layer for object $i$ at frame $t$ depends on the following variables: $(a)$ the features $f_t$ obtained from the encoder from frame $t$, $(b)$ the preceding $k-1$-th ConvLSTM layer, $(c)$ the hidden state representation from the previous object $i-1$ at the same frame $t$, i.e. $h_{t,i-1,k}$, which will be referred to as the \emph{spatial hidden state}, $(d)$ the hidden state representation representation from the same object $i$ at the previous frame $t-1$, i.e. $h_{t-1,i,k}$, which will be referred to as the \emph{temporal hidden state}, and $(e)$ the object segmentation prediction mask $S_{t-1,i}$ of the object $i$ at the previous frame $t-1$:

\begin{align}
\label{eq:ith_convlstm_mask}
&h_{input} =  \;[\; B_{2}(h_{t,i,k-1}) \; | \; f'_{t,k} \; | \; S_{t-1,i} \;]\\
&h_{state} = \;[\; h_{t,i-1,k} \; | \; h_{t-1,i,k}  \;] \;\\
\label{eq:ith_convlstm}
&h_{t,i,k} = \mathrm{ConvLSTM_{k}}( \; h_{input} \; , \; h_{state} \; )
\end{align}
where $B_{2}$ is the bilinear upsampling operator by a factor of 2 and $f'_{t,k}$ is the result of projecting $f_{t,k}$ to have lower dimensionality via a convolutional layer.

Equation~\ref{eq:ith_convlstm} is applied in chain for $k \in \{1,...,n_b\}$, being $n_b$ the number of convolutional blocks in the encoder. $h_{t,i,0}$ is obtained by considering

\begin{equation*}
    h_{input} =  \;[\; f'_{t,0} \; | \; S_{t-1,i} \;]
\end{equation*}
and for the first object, $h_{state}$ is obtained as follows:

\begin{equation*}
    h_{state} = \;[\; Z \; | \; h_{t-1,i,k}  \;] \;\\
\end{equation*}
where $Z$ is a zero matrix that represents that there is no previous spatial hidden state for this object.

In Section~\ref{sec:experiments}, an ablation study will be performed in order to analyze the importance of spatial and temporal recurrence in the decoder for the VOS task.

%% file: sections/4_experiments.tex
\section{Experiments}
\label{sec:experiments}

The experiments are carried out for two different tasks of the VOS: the one-shot and the zero-shot. In both cases, we analyze how important the spatial and the temporal hidden states are. Thus, we consider three different options: $(i)$ spatial model (temporal recurrence is not used), $(ii)$ temporal model (spatial recurrence is not used), and $(iii)$ spatio-temporal model (both spatial and temporal recurrence are used). In the one-shot VOS, since the masks for the objects at the first frame are given, the decoder always considers the mask $S_{t-1,i}$ from the previous frame when computing $h_{input}$ (see Eq.~\ref{eq:ith_convlstm_mask}). On the other hand, in the zero-shot VOS, $S_{t-1,i}$ is not used since no ground truth masks are given.

The experiments are performed in the two most recent VOS benchmarks: YouTube-VOS \cite{xu2018youtube-benchmark} and DAVIS-2017 \cite{Pont-Tuset_arXiv_2017}. YouTube-VOS consists of 3,471 videos in the training set and 474 videos in the validation set, being the largest video object segmentation benchmark. The training set includes 65 unique object categories which are regarded as seen categories. In the validation set,  there are 91 unique object categories, which include all the seen categories and 26 unseen categories. On the other hand, DAVIS-2017 consists of 60 videos in the training set, 30 videos in the validation set and 30 videos in the test-dev set. Evaluation is performed on the YouTube-VOS validation set and on the DAVIS-2017 test-dev set. Both YouTube-VOS and DAVIS-2017 videos include multiple objects and have a similar duration in time (3-6 seconds).

The experiments are evaluated using the usual evaluation measures for VOS: $(i)$ the region similarity $J$, and $(ii)$ the contour accuracy $F$. In YouTube-VOS, each of these measures is split into two different measures, depending on whether the categories have already been seen by the model ($J_{seen}$ and $F_{seen}$), i.e. these categories are included in the training set, or the model has never seen these categories ($J_{unseen}$ and $F_{unseen}$).

\subsection{One-shot video object segmentation}


One-shot VOS consists in segmenting the objects from a video given the objects masks from the first frame. Since the initial masks are given, the experiments have been performed including the mask of the previous frame as one additional input channel in the ConvLSTMs from our decoder. 

\begin{figure}
    \centering
    \includegraphics[width=0.95\columnwidth]{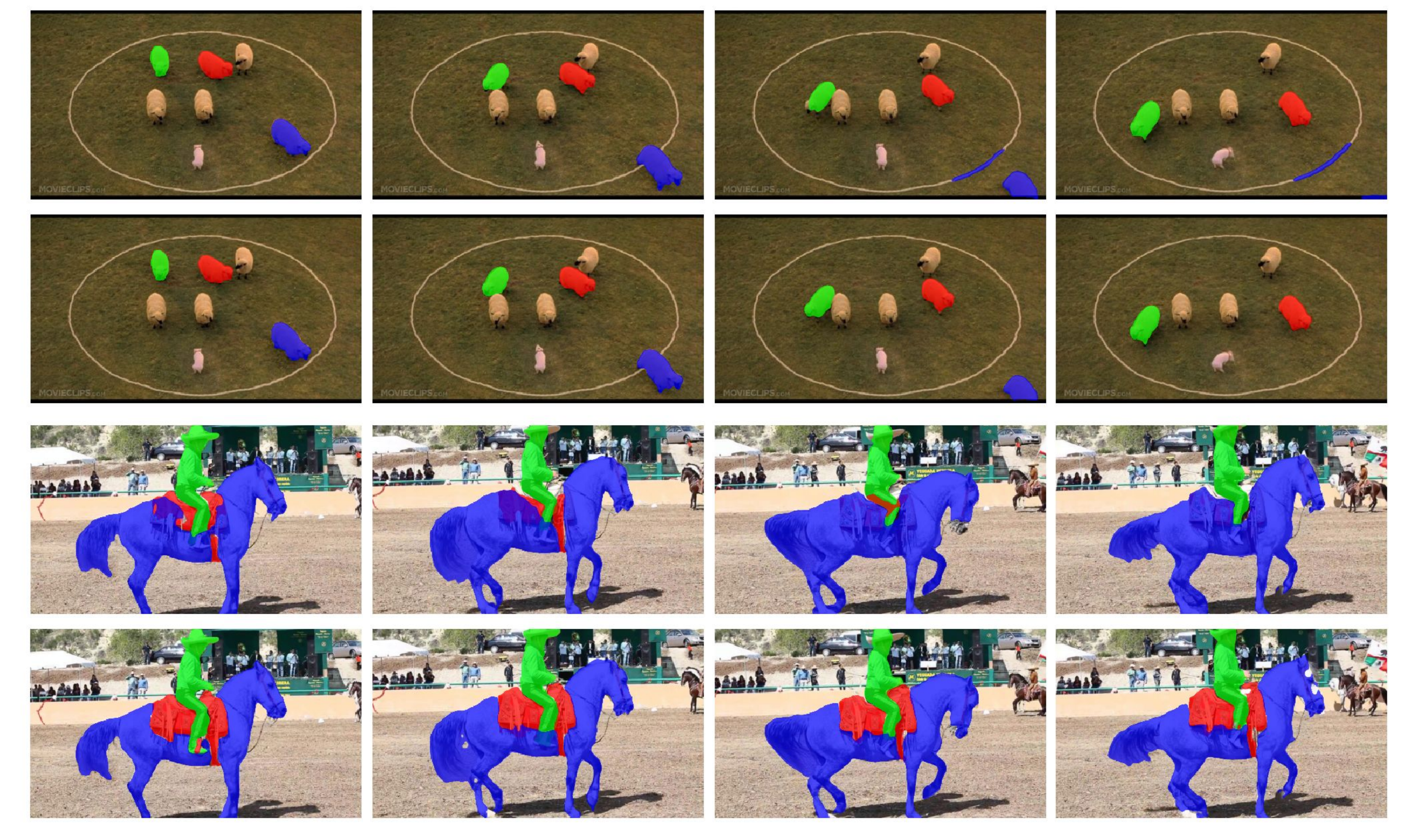}
    \caption{Qualitative results comparing spatial (rows 1,3) and spatio-temporal (rows 2,4) models.}
    \label{fig:spatial-spatiotemporal-comparison}
\end{figure}

\textbf{YouTube-VOS benchmark} Table~\ref{tab:oneshot_youtube_ablation} shows the results obtained in YouTube-VOS validation set for different configurations: spatial (RVOS-Mask-S), temporal (RVOS-Mask-T) and spatio-temporal (RVOS-Mask-ST). All models from this ablation study have been trained using a 80\%-20\% split of the training set. 
We can see that the spatio-temporal model improves both the region similarity $J$ and contour accuracy $F$ for seen and unseen categories over the spatial and temporal models. Figure~\ref{fig:spatial-spatiotemporal-comparison} shows some qualitative results comparing the spatial and the spatio-temporal models, where we can see that the RVOS-Mask-ST preserves better the segmentation of the objects along the time.

\begin{figure}
    \centering
    \includegraphics[width=0.95\columnwidth]{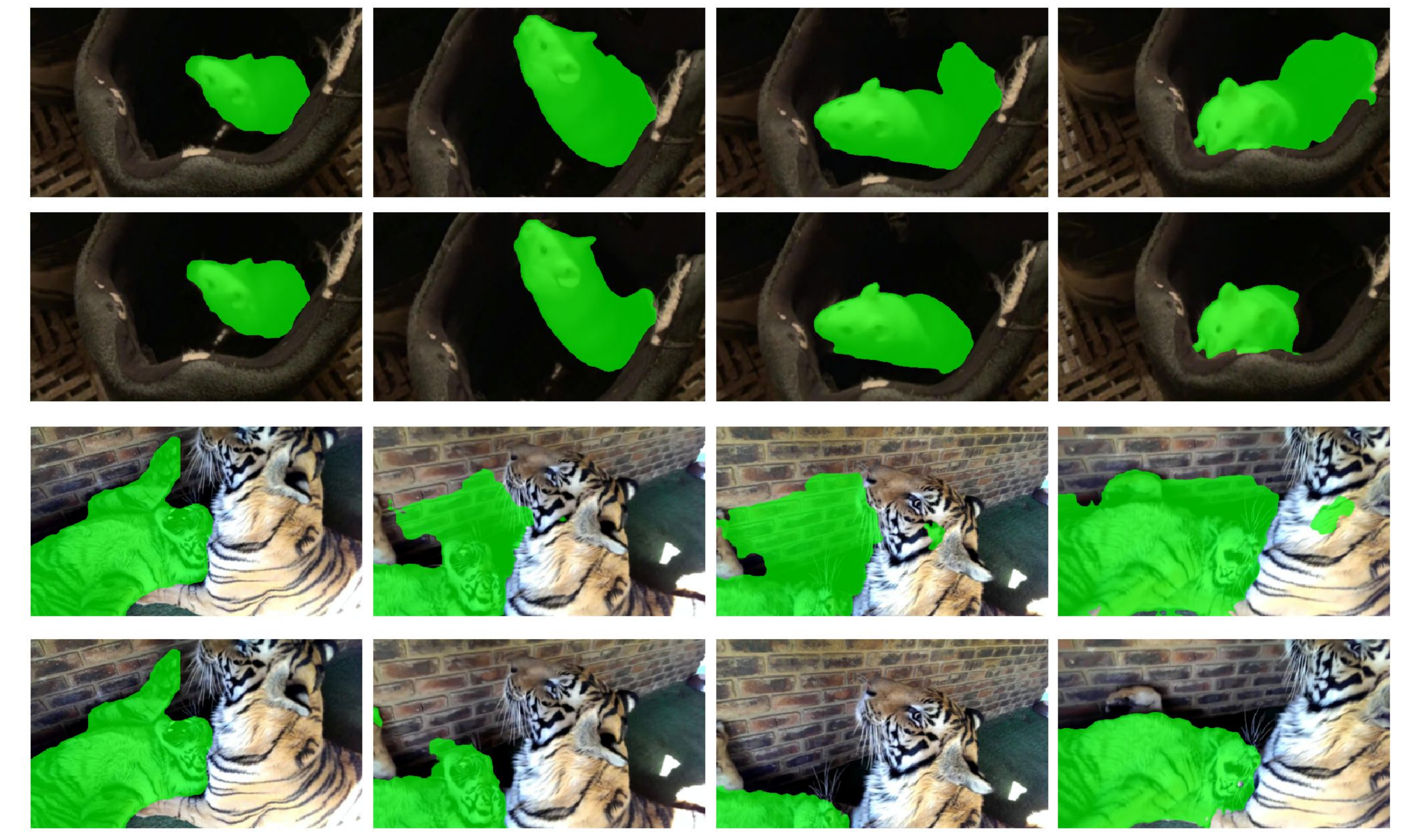}
    \caption{Qualitative results comparing training with ground truth masks (rows 1,3) and training with inferred masks (rows 2,4).}
    \label{fig:gtmask-inferencemask-comparison}
\end{figure}

Furthermore, we have also considered fine-tuning the models some additional epochs using the inferred mask from the previous frame $\hat{S}_{t-1,i}$, instead of using the ground truth mask $S_{t-1,i}$. This way, the model can learn how to fix some errors that may occur in inference. 
In Table~\ref{tab:oneshot_youtube_ablation}, we can see that this model (RVOS-Mask-ST+) is more robust and outperforms the model trained only with the ground truth masks. Figure~\ref{fig:gtmask-inferencemask-comparison} shows some qualitative results comparing the model trained with the ground truth mask and the model trained with the inferred mask.

\begin{table}[]
\centering
\begin{tabular}{@{}lcccc@{}}
\toprule
\multicolumn{1}{c}{}  &  \multicolumn{4}{c}{\textbf{YouTube-VOS one-shot}} \\
    & $J_{seen}$    & $J_{unseen}$   &    $F_{seen}$   & $F_{unseen}$  \\  
\midrule
RVOS-Mask-S                              & 54.7    & 37.3    & 57.4 & 42.4 \\
RVOS-Mask-T                              & 59.9    & 39.2    & 63.1 & 45.6 \\
RVOS-Mask-ST                             & 60.8    & \textbf{44.6}    & 63.7 & 50.3 \\
RVOS-Mask-ST+                            & \textbf{63.1}    & 44.5    & \textbf{67.1} & \textbf{50.4} \\
\bottomrule
\end{tabular}
\caption{Ablation study about spatial and temporal recurrence in the decoder for one-shot VOS in YouTube-VOS dataset. Models have been trained using 80\%-20\% partition of the training set and evaluated on the validation set. + means that the model has been trained using the inferred masks.}
\label{tab:oneshot_youtube_ablation}
\end{table}

Once stated that the spatio-temporal model is the model that gives the best performance, we have trained the model using the whole YouTube-VOS training set to compare it with other state-of-the-art techniques (see Table~\ref{tab:oneshot_youtube_comparison}). Our proposed spatio-temporal model (RVOS-Mask-ST+) has comparable results with respect to S2S w/o OL \cite{xu2018youtube-benchmark}, with a slightly worse performance in region similarity $J$ but with a slightly better performance in contour accuracy $F$. Our model outperforms the rest of state-of-the-art techniques \cite{caelles2017one,perazzi2017learning,yang2018efficient,voigtlaender2017online} for the \emph{seen} categories. It is OSVOS \cite{caelles2017one} the one that gives the best performance for the \emph{unseen} categories. However, note that the comparison of S2S without online learning \cite{xu2018youtube-benchmark} and our proposed model with respect to OSVOS \cite{caelles2017one}, OnAVOS \cite{voigtlaender2017online} and MaskTrack \cite{perazzi2017learning} is not fair for $J_{unseen}$ and $F_{unseen}$ because OSVOS, OnAVOS and MaskTrack models are finetuned using the annotations of the first frames from the validation set, i.e. they use online learning. Therefore, \emph{unseen} categories should not be considered as such since the model has already seen them.

\begin{table}[]
\centering
\resizebox{\columnwidth}{!}{
\begin{tabular}{@{}lccccc@{}}
\toprule
\multicolumn{2}{c}{}  &  \multicolumn{4}{c}{\textbf{YouTube-VOS one-shot}} \\
  & OL  & $J_{seen}$    & $J_{unseen}$   &    $F_{seen}$   & $F_{unseen}$  \\  
\midrule
OSVOS \cite{caelles2017one}        & \cmark         & 59.8 & \textbf{54.2} & 60.5 & \textbf{60.7} \\
MaskTrack  \cite{perazzi2017learning} & \cmark            & 59.9 & 45.0 & 59.5 & 47.9 \\
OnAVOS \cite{voigtlaender2017online}    & \cmark             & \textbf{60.1} & 46.6 & \textbf{62.7} & 51.4 \\
\midrule
OSMN \cite{yang2018efficient}    & \xmark                & 60.0 & 40.6 & 60.1 & 44.0 \\
S2S w/o OL \cite{xu2018youtube-benchmark}  & \xmark     & \textbf{66.7}    & \textbf{48.2}    & 65.5 & 50.3 \\
RVOS-Mask-ST+       & \xmark      & 63.6    & 45.5    & \textbf{67.2} & \textbf{51.0} \\
\bottomrule
\end{tabular}}
\caption{Comparison against state of the art VOS techniques for one-shot VOS on YouTube-VOS validation set. OL refers to online learning. The table is split in two parts, depending on whether the techniques use online learning or not.}
\label{tab:oneshot_youtube_comparison}
\end{table}
Table~\ref{tab:num-instances-analysis} shows the results on the region similarity $J$ and the contour accuracy $F$ depending on the number of instances in the videos. We can see that the fewer the objects to segment, the easier the task, obtaining the best results for sequences where only one or two objects are annotated. 

\begin{table}[]
    \centering
    \begin{tabular}{lccccc}
\toprule
\multicolumn{1}{c}{}  &  \multicolumn{5}{c}{\textbf{Number of instances (YouTube-VOS)}} \\
& 1 & 2 & 3 & 4 & 5 \\
\midrule
$J$ mean & 78.2 & 62.8 & 50.7 & 50.2 & 56.3 \\
$F$ mean & 75.5 & 67.6 & 56.1 & 62.3 & 66.4 \\
\bottomrule
    \end{tabular}
    \caption{Analysis of our proposed model RVOS-Mask-ST+ depending on the number of instances in one-shot VOS.}
    \vspace{-3mm}

    \label{tab:num-instances-analysis}
\end{table}

\begin{figure*}
    \centering
    \includegraphics[width=\textwidth]{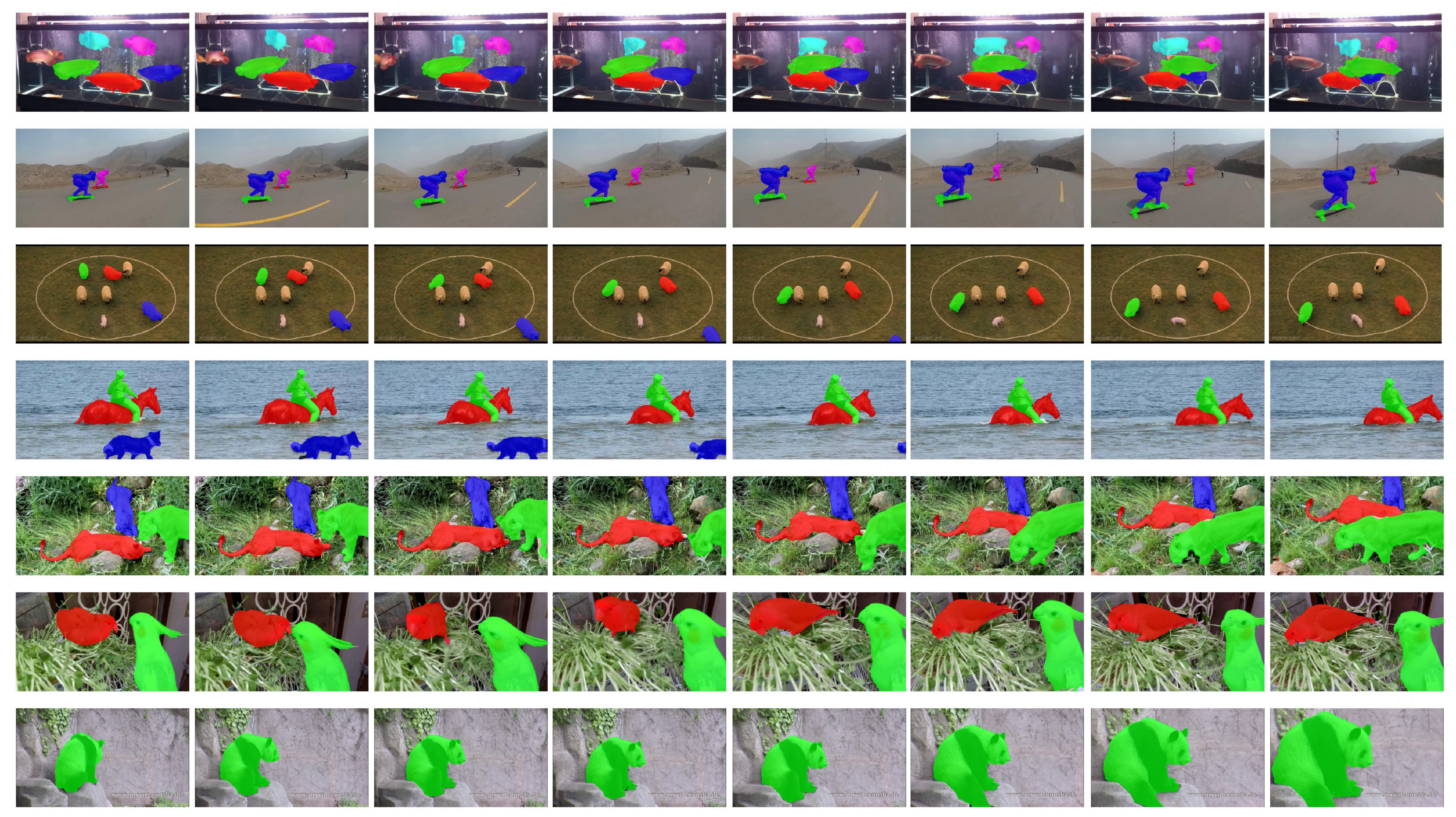}
    \caption{Qualitative results for one-shot video object segmentation on YouTube-VOS with multiple instances.}
    \label{fig:youtube-oneshot-qualitative-results}
\end{figure*}

Figure~\ref{fig:youtube-oneshot-qualitative-results} shows some qualitative results of our spatio-temporal model for different sequences from YouTube-VOS validation set. It includes examples with different number of instances. Note that the instances have been properly segmented although there are different instances of the same category in the sequence (fishes, sheeps, people, leopards or birds) or there are some instances that disappear from the sequence (one sheep in third row or the dog in fourth row). 

\textbf{DAVIS-2017 benchmark} Our pretrained model RVOS-Mask-ST+ in YouTube-VOS has been tested on a different benchmark: DAVIS-2017. As it can be seen in Table~\ref{tab:oneshot_davis_comparison}, when the pretrained model is directly applied to DAVIS-2017, RVOS-Mask-ST+ (pre) outperforms the rest of state-of-the-art techniques that do not make use of online learning, i.e. OSMN~\cite{yang2018efficient} and FAVOS~\cite{Cheng_favos_2018}. Furthermore, when the model is further finetuned for the DAVIS-2017 training set, RVOS-Mask-ST+ (ft) outperforms some techniques as OSVOS \cite{caelles2017one}, which is among the techniques that make use of online learning. Note that online learning requires finetuning the model at test time.

\begin{table}[]
\centering
\begin{tabular}{@{}lccc@{}}
\toprule
\multicolumn{2}{c}{}  &  \multicolumn{2}{c}{\textbf{DAVIS-2017 one-shot}} \\
  & OL  & $J$    & $F$  \\  
\midrule
OSVOS \cite{caelles2017one}        & \cmark         & 47.0 & 54.8 \\
OnAVOS \cite{voigtlaender2017online}    & \cmark             & 49.9 & 55.7 \\
OSVOS-S \cite{maninis2018video} & \cmark             & 52.9 & 62.1 \\
CINM \cite{bao2018cnn} & \cmark             & \textbf{64.5} & \textbf{70.5} \\
\midrule
OSMN \cite{yang2018efficient} & \xmark             & 37.7 & 44.9 \\
FAVOS \cite{Cheng_favos_2018} & \xmark             & 42.9 & 44.2 \\
RVOS-Mask-ST+ (pre)       & \xmark      & 46.4    & 50.6   \\
RVOS-Mask-ST+ (ft)       & \xmark      & \textbf{48.0}    & \textbf{52.6}   \\
\bottomrule
\end{tabular}
\caption{Comparison against state of the art VOS techniques for one-shot VOS on DAVIS-2017 test-dev set. OL refers to online learning. The model RVOS-Mask-ST+(pre) is the one trained on Youtube-VOS, and the model RVOS-Mask-ST+ (ft) is after finetuning the model for DAVIS-2017. The table is split in two parts, depending on whether the techniques use online learning or not.}
\label{tab:oneshot_davis_comparison}
\end{table}

\begin{figure}
    \centering
    \includegraphics[width=0.95\columnwidth]{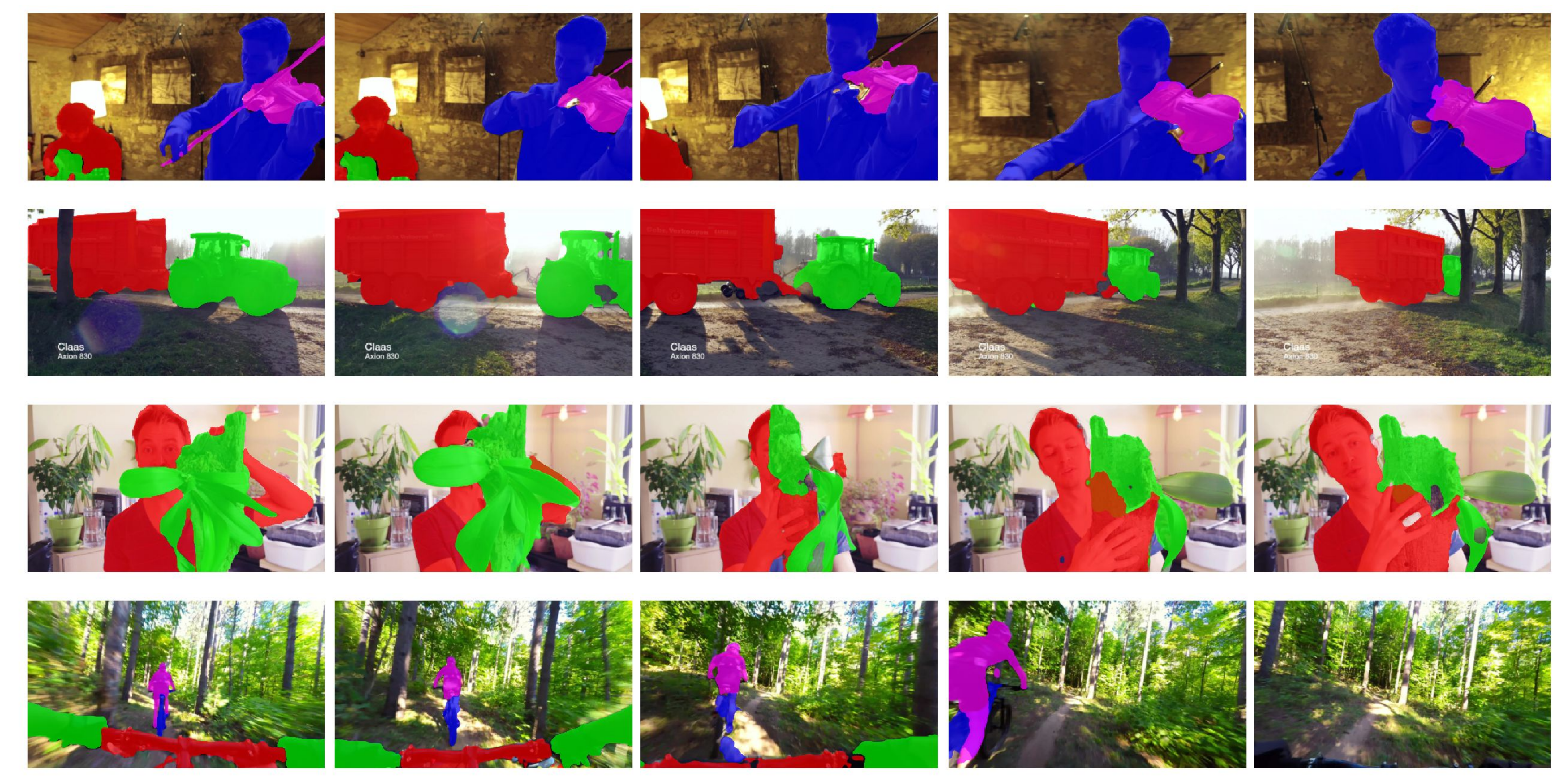}
    \caption{Qualitative results for one-shot on DAVIS-2017 test-dev.}
    \label{fig:one-shot-davis}
\end{figure}

Figure~\ref{fig:one-shot-davis} shows some qualitative results obtained for DAVIS-2017 one-shot VOS. As depicted in some qualitative results for YouTube-VOS, RVOS-Mask-ST+ (ft) is also able to deal with objects that disappear from the sequence.

\subsection{Zero-shot video object segmentation}

Zero-shot VOS consists in segmenting the objects from a video without having any prior knowledge about which objects have to be segmented, i.e. no object masks are provided. This task is more complex that the one-shot VOS since the model has to detect and segment the objects appearing in the video. 

\begin{figure}
    \centering
    \includegraphics[width=0.95\columnwidth]{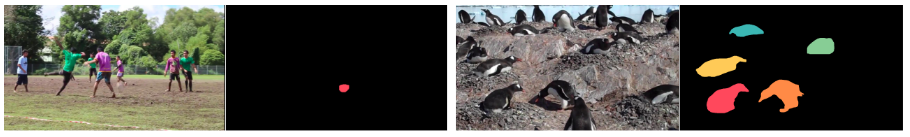}
    \caption{Missing object annotations may suppose a problem for zero-shot video object segmentation. 
    }
    \label{fig:missing-annotations}
\end{figure}

Nowadays, to our best knowledge, there is no benchmark specially designed for zero-shot VOS. Although YouTube-VOS and DAVIS benchmarks can be used for training and evaluating the models without using the annotations given at the first frame, both benchmarks have the limitation that not all objects appearing in the video are annotated. Specifically, in YouTube-VOS, there are up to 5 object instances annotated per video. This makes sense when the objects to segment are given (as done in one-shot VOS), but it may be a problem for zero-shot VOS since the model could be segmenting correctly objects that have not been annotated in the dataset. Figure~\ref{fig:missing-annotations} shows a couple of examples where there are some missing object annotations. 

Despite the problem stated before about missing object annotations, we have trained our model for the zero-shot VOS problem using the object annotations available in these datasets. To minimize the effect of segmenting objects that are not annotated and missing the ones that are annotated, we allow our system to segment up to 10 object instances along the sequence, expecting that the up to 5 annotated objects are among the predicted ones. During training, each annotated object is uniquely assigned to one predicted object to compute the loss. Therefore, predicted objects which have not been assigned do not result in any loss penalization. However, the bad prediction of any annotated object is considered by the loss. Analogously, in inference, in order to evaluate our results for zero-shot video object segmentation, the masks provided for the first frame in one-shot VOS are used to select which predicted instances are selected for evaluation. Note that the assignment is only performed at the first frame and the predicted segmentation masks considered for the rest of the frames are the corresponding ones. 

\textbf{YouTube-VOS benchmark} Table~\ref{tab:zeroshot_youtube} shows the results obtained on YouTube-VOS validation set for the zero-shot VOS problem. As stated for the one-shot VOS problem, the spatio-temporal model (RVOS-ST) also outperforms both spatial (RVOS-S) and temporal (RVOS-T) models. 

\begin{table}[]
\centering
\begin{tabular}{@{}lcccc@{}}
\toprule
\multicolumn{1}{c}{}  &  \multicolumn{4}{c}{\textbf{YouTube-VOS zero-shot}} \\
    & $J_{seen}$  &  $J_{unseen}$ &    $F_{seen}$  & $F_{unseen}$  \\  
\midrule
RVOS-S                              & 40.8    & 19.9    & 43.9 & 23.2 \\
RVOS-T                              & 37.1    & 20.2    & 38.7 & 21.6 \\
RVOS-ST                             & \textbf{44.7}    & \textbf{21.2}    & \textbf{45.0} & \textbf{23.9} \\
\bottomrule
\end{tabular}
\caption{Ablation study about spatial and temporal recurrence in the decoder for zero-shot VOS in YouTube-VOS dataset. Our models have been trained using 80\%-20\% partition of the training set and evaluated on the validation set.}
\label{tab:zeroshot_youtube}
\end{table}

\begin{figure}
    \centering
    \includegraphics[width=0.92\columnwidth]{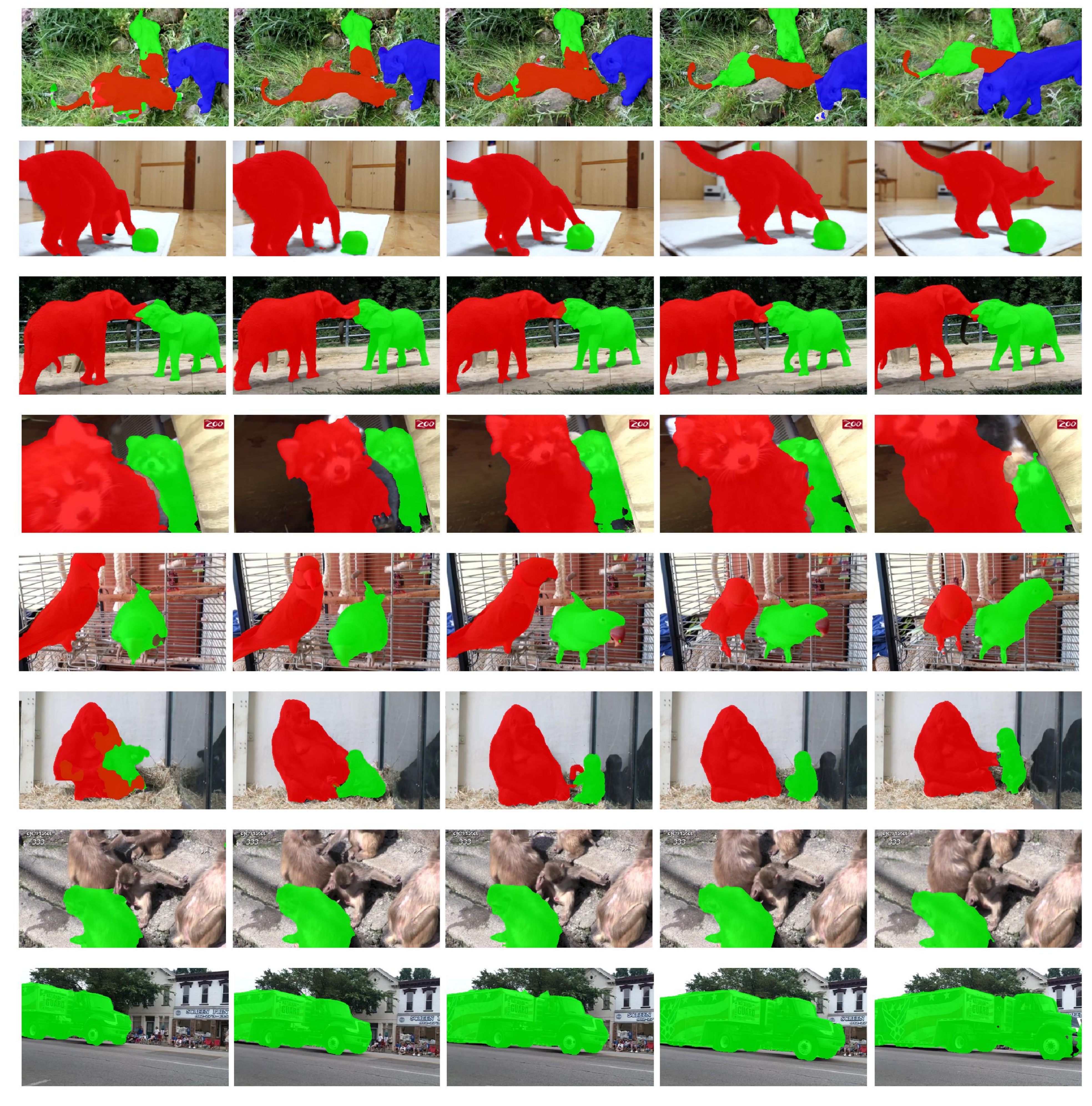}
    \caption{Qualitative results for zero-shot video object segmentation on YouTube-VOS with multiple instances.}
    \label{fig:youtube-zeroshot-qualitative-results}
\end{figure}

Figure~\ref{fig:youtube-zeroshot-qualitative-results} shows some qualitative results for zero-shot VOS in YouTube-VOS validation set. Note that the masks are not provided and the model has to discover the objects to be segmented. We can see that in many cases our spatio-temporal model is temporal consistent although the sequence contains different instances of the same category.

\textbf{DAVIS-2017 benchmark} To our best knowledge, there are no published results for this task in DAVIS-2017 to be compared. The zero-shot VOS has only been considered for DAVIS-2016, where some unsupervised techniques have been applied. However, in DAVIS-2016, there is only a single object annotated for sequence, which could be considered as a foreground-background video segmentation problem and not as a multi-object video object segmentation. Our pretrained model RVOS-ST on Youtube-VOS for zero-shot, when it is directly applied to DAVIS-2017, obtains a mean region similarity $J=21.7$ and a mean contour accuracy $F=27.3$. When the pretrained model is finetuned for the DAVIS-2017 trainval set achieves a slightly better performance, with $J=23.0$ and $F=29.9$.

\begin{figure}
    \centering
    \includegraphics[width=0.92\columnwidth]{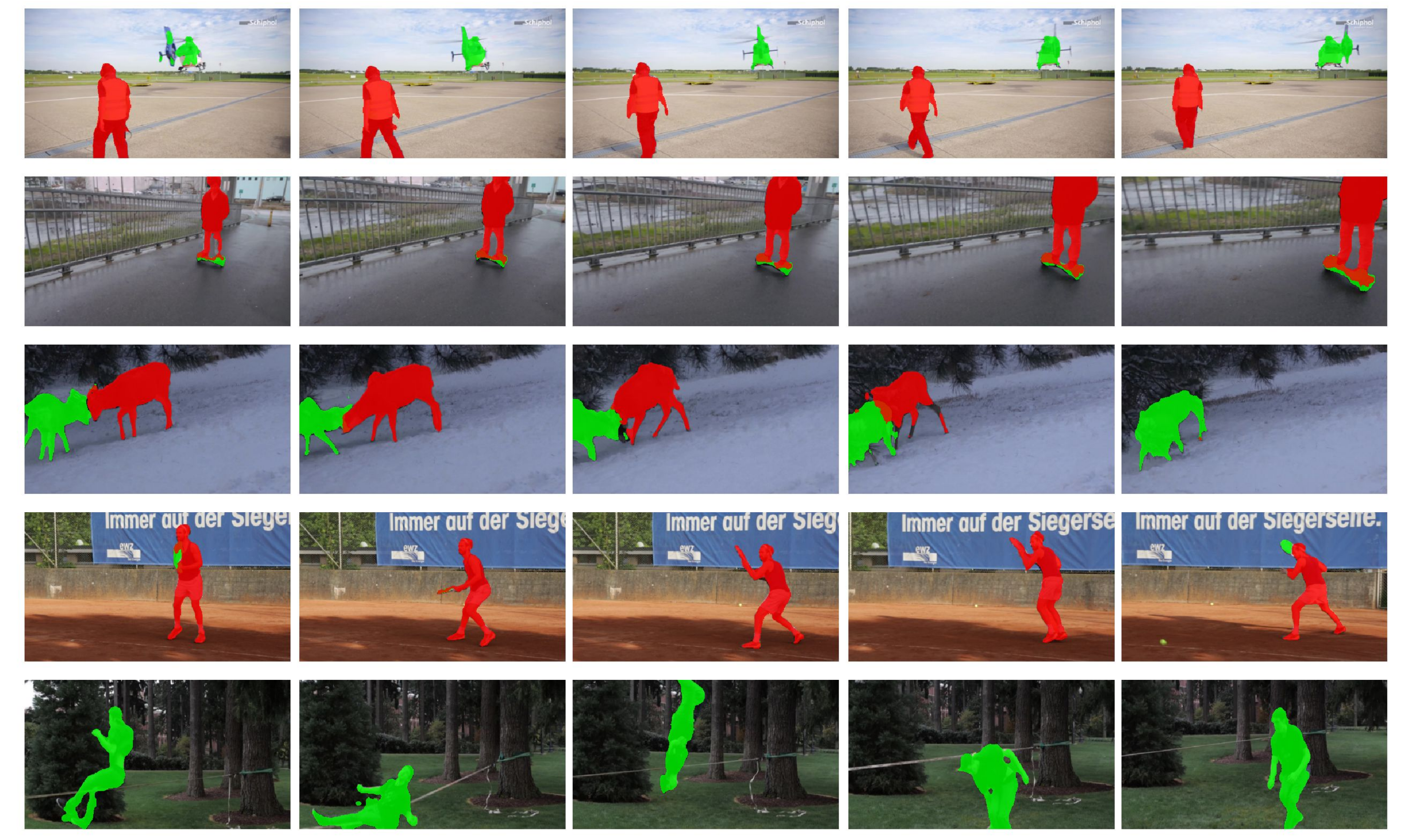}
    \caption{Qualitative results for zero-shot video object segmentation on DAVIS-2017 with multiple instances.}
    \label{fig:davis-zeroshot-qualitative-results}
\end{figure}

Although the model has been trained on a large video dataset as Youtube-VOS, there are some sequences where the object instances have not been segmented from the beginning. The low performance for zero-shot VOS in DAVIS-2017 ($J=23.0$) can be explained due to the bad performance also in YouTube-VOS for the \emph{unseen} categories ($J_{unseen}=21.2$). Therefore, while the model is able to segment properly categories which are included among the YouTube-VOS training set categories, e.g. persons or animals, the model fails when trying to segment an object which has not been seen before. Note that it is specially for these cases when online learning becomes relevant, since it allows to finetune the model by leveraging the object mask given at the first frame for the one-shot VOS problem. Figure~\ref{fig:davis-zeroshot-qualitative-results} shows some qualitative results for the DAVIS-2017 test-dev set when no object mask is provided where our RVOS-ST model has been able to segment the multiple object instances appearing in the sequences.

\subsection{Runtime analysis and training details}

\textbf{Runtime analysis} Our model (RVOS) is the fastest method amongst all while achieving comparable segmentation quality with respect to state-of-the-art as seen previously in Tables \ref{tab:oneshot_youtube_comparison} and \ref{tab:oneshot_davis_comparison}. The inference time for RVOS is 44ms per frame with a GPU P100 and 67ms per frame with a GPU K80. Methods not using online learning (including ours) are two orders of magnitude faster than techniques using online learning. Inference times for OSMN \cite{yang2018efficient} (140ms) and S2S \cite{xu2018youtube-benchmark} (160ms) have been obtained from their respective papers. For a fair comparison, we also compute runtimes for OSMN \cite{yang2018efficient} in our machines (K80 and P100) using their public implementation (no publicly available code was found for \cite{xu2018youtube-benchmark}). 
We measured better runtimes for OSMN than those reported in \cite{yang2018efficient}, but RVOS is still faster in all cases (e.g. 65ms vs. 44ms on a P100, respectively).
To the best of our knowledge, our method is the first to share the encoder forward pass for all the objects in a frame, which explains its fast overall runtime. 


\textbf{Training details} The original RGB frames and annotations have been resized to 256$\times$448 in order to have a fair comparison with S2S \cite{xu2018youtube} in terms of image resolution. In training, due to memory restrictions, each training mini-batch is composed with 4 clips of 5 consecutive frames. However, in inference, the hidden state is propagated along the whole video. Adam optimizer is used to train our network and the initial learning rate is set to $10^{-6}$. Our model has been trained for 20 epochs using the previous ground truth mask and 20 epochs using the previous inferred mask in a single GPU with 12GB RAM, taking about 2 days.

%% file: sections/5_conclusion.tex
\section{Conclusions}

In this work we have presented a fully end-to-end trainable model for multiple objects in video object segmentation (VOS) with a recurrence module based on spatial and temporal domains. The model has been designed for both one-shot and zero-shot VOS and tested on YouTube-VOS and DAVIS-2017 benchmarks. 

The experiments performed show that the model trained with spatio-temporal recurrence improves the models that only consider the spatial or the temporal domain. We give the first results for zero-shot VOS on both benchmarks and we also outperform state-of-the-art techniques that do not make use of online learning for one-shot VOS on them. 

The code is available in our project website\footnote{{\href{https://imatge-upc.github.io/rvos/}{https://imatge-upc.github.io/rvos/}}}.

\label{sec:conclusion}

%% file: sections/6_acks.tex
\section*{Acknowledgements}

This research was supported by the Spanish Ministry of Economy and Competitiveness and the European Regional Development Fund (TIN2015-66951-C2-2-R, TIN2015-65316-P \& TEC2016-75976-R), the BSC-CNS Severo Ochoa SEV-2015-0493 and LaCaixa-Severo Ochoa International Doctoral Fellowship programs, the 2017 SGR 1414 and the Industrial Doctorates 2017-DI-064 \& 2017-DI-028 from the Government of Catalonia.